\documentclass[journal]{IEEEtran}
\usepackage{amsmath,amsfonts}
\usepackage{algorithmic}
\usepackage{array}
\usepackage{textcomp}
\usepackage{stfloats}
\usepackage[ruled,linesnumbered]{algorithm2e}
\usepackage{subfig}
\usepackage{verbatim}
\usepackage{graphicx}

\def\BibTeX{{\rm B\kern-.05em{\sc i\kern-.025em b}\kern-.08em T\kern-.1667em\lower.7ex\hbox{E}\kern-.125emX}}
\usepackage{balance}

\begin{document}

\title{Real-world Troublemaker: A 5G Cloud-controlled Track Testing Framework for Automated Driving Systems in Safety-critical Interaction Scenarios}
\author{Xinrui Zhang, Lu Xiong, Peizhi Zhang, Junpeng Huang, and Yining Ma
}

\maketitle

\begin{abstract}
Track testing plays a critical role in the safety evaluation of autonomous driving systems (ADS), as it provides a real-world interaction environment. However, the inflexibility in motion control of object targets and the absence of intelligent interactive testing methods often result in pre-fixed and limited testing scenarios. To address these limitations, we propose a novel 5G cloud-controlled track testing framework, Real-world Troublemaker. This framework overcomes the rigidity of traditional pre-programmed control by leveraging 5G cloud-controlled object targets integrated with the Internet of Things (IoT) and vehicle teleoperation technologies. Unlike conventional testing methods that rely on pre-set conditions, we propose a dynamic game strategy based on a quadratic risk interaction utility function, facilitating intelligent interactions with the vehicle under test (VUT) and creating a more realistic and dynamic interaction environment. The proposed framework has been successfully implemented at the Tongji University Intelligent Connected Vehicle Evaluation Base. Field test results demonstrate that Troublemaker can perform dynamic interactive testing of ADS accurately and effectively. Compared to traditional methods, Troublemaker improves scenario reproduction accuracy by 65.2\%, increases the diversity of interaction strategies by approximately 9.2 times, and enhances exposure frequency of safety-critical scenarios by 3.5 times in unprotected left-turn scenarios.

\end{abstract}


\section{Introduction}
\IEEEPARstart{T}{he} safety of automated driving systems (ADS) must be ensured prior to their practical implementation, which requires a well-established testing framework \cite{bai2024ar}. Existing test standards, such as ISO 26262 \cite{iso201126262}, UN R157 \cite{no2021157}, and UN R171 \cite{UNR171}, are not sufficient to comprehensively evaluate ADS. According to the driving automation levels defined by SAE J3016 from SAE International, a high-level ADS (i.e., Level 3 or higher) is expected to perform driving tasks independently and autonomously, with the driver no longer retaining continuous control over vehicle movement \cite{SAE_202104}. While ADS has already been deployed in various countries and regions, numerous ADS traffic incidents highlight that safety testing for high-level ADS remains a critical technical challenge. In comparison to traditional vehicles and advanced driver assistance systems (ADAS), high-level ADS testing faces significant transformations and challenges, particularly in terms of both test subjects and requirements.

In terms of test subjects, ADS has become more complex and intelligent, with the testing paradigm evolving from the traditional binary human-vehicle approach to a tightly coupled ``vehicle-road-environment-task" system. From a testing entity perspective, ADS is no longer standalone like traditional vehicles but is now embedded within the broader transportation and traffic ecosystem \cite{szalay2019multi}. With the rise of machine learning and end-to-end algorithms in high-level ADS technologies, the testing challenges have intensified. These advanced vehicles can outsmart predefined tests. As such, relying solely on dummy test equipment and modular testing methods is no longer sufficient to test these more intelligent and sophisticated systems.
\IEEEpubidadjcol

In terms of testing requirements, the primary focus of ADS has shifted from vehicle mechanical performance to driving performance in high-interaction scenarios. The ultimate goal of ADS testing is to validate the performance of the system under real-world conditions, ensuring that ADS can outperform the average human driver in terms of safety \cite{sun2021scenario}. The literature suggests that, to demonstrate that autonomous vehicles have a lower accident rate than human drivers in complex, real-world traffic environments, at least 17.7 billion kilometers of public road testing are required \cite{kalra2016driving}. To fully capture the complexity and variability of these environments, the variables that define traffic scenarios are inherently complex, with interactive behaviors being random and diverse, which can lead to the curse of dimensionality \cite{mullins2018adaptive}. Moreover, traditional testing methods may fail to address safety-critical scenarios, particularly due to the long-tail problem.

Currently, ADS assessment is performed mainly in simulation testing, track testing, and public road testing. Simulation testing effectively addresses the issue of mileage testing, but modeling precise and complex interaction scenarios remains a significant challenge \cite{feng2020testing1}. Public road testing, while the most representative of real-world conditions, carries substantial and unacceptable risks, along with potential losses in the event of failures \cite{li2018artificial}. Additionally, public road testing is often inefficient, as many scenarios encountered are non-safety-critical and repetitive, which offers little value for the safety validation of ADS \cite{duan2024digital}. Compared with simulation and public road testing, track testing has its unique advantages. First, test tracks provide a realistic yet manageable environment by utilizing soft targets. Second, it enables the testing of interaction scenarios that closely simulate real traffic conditions \cite{batsch2021taxonomy}. In this setting, the vehicle under test (VUT) can be evaluated on roads with controlled interactions involving other actors, such as global vehicle targets and pedestrian targets, for crash testing. Third, dozens of test tracks are constructed around the world \cite{emami2022review}.

Focused on ADS testing requirements, extensive research has been conducted around the test equipment and interaction scenarios test methods for track testing. On the aspect of test equipment, the definition of object targets for track testing systems has led to the establishment of corresponding standards. The ISO 19206 standard defines the test devices for targets and, through reasonable control methods, enables the actual ADS to be tested in traffic environments designed to simulate specific challenging driving conditions. However, the existing standards primarily focus on the physical characteristics of the test targets, such as size and reflective properties, while the control of these targets is still mainly oriented toward ADAS testing. On the aspect of interaction scenario test methods, a widely accepted classification has been established, which includes the following:
\subsubsection{Predefined Triggered Scenarios} Test scenarios are often selected based on a variety of parameter combinations. The advantage of this method lies in its repeatability, reliability, and ability to complete tests within a reasonable timeframe. However, it remains unclear how the selected test scenarios correlate with real-world conditions, particularly when human interaction is involved \cite{leblanc2014tradeoffs}. Furthermore, because all test scenarios are fixed and predefined, AVs can be tuned to achieve great performance in these tests, but their behaviors under broader conditions are not adequately assessed \cite{batsch2021taxonomy}. 
\subsubsection{Worst-case Scenario Evaluation (WCSE)} This methodology involves selecting the most challenging scenarios for testing. A limitation of this approach is its failure to consider the probability of extreme scenarios occurring, which may lead to the creation of unrealistic scenarios that do not accurately represent real traffic conditions \cite{sun2021scenario}. 
\subsubsection{Search-based Methods} It is widely used due to its effectiveness in generating challenging scenarios. The mainstream approach to generating these scenarios involves constructing key parameter state spaces and searching for specific scenarios within a logical scenario space \cite{li2024hazardous}. Various strategies such as adaptive sampling \cite{gong2023adaptive} \cite{sun2021adaptive}, optimization \cite{abeysirigoonawardena2019generating} \cite{zhu2021hazardous}, evolutionary algorithms \cite{klischat2019generating}, and reinforcement learning \cite{liu2024safety} are used for risk scenario search. However, these methods are difficult to characterize the interactivity of traffic driving behavior, often resulting in many non-safety-critical and trivial scenarios. Game theory has the advantage of capturing the interdependencies of actions between decision-makers and has been widely used for modeling behavior in interactive scenarios. The level-k game framework builds differentiated interaction strategies based on varying styles for verification and validation of autonomous vehicle control systems \cite{li2017game}. Li et al. \cite{li2023two} used k-means algorithms for driving style classification and designed overtaking strategies with the game framework based on different styles. However, the interactive scenarios modeled in these game-theoretic approaches focus more on human-like behavior and lack a model that describes interaction strategies from the perspective of interaction risks. This makes it difficult to construct the required interactive scenarios for testing purposes.

The state-of-the-art method in the field of ADS track testing is accelerated testing. Research focused on improving the efficiency of search-based testing methods. A risk scenario is defined as the combination of maneuver challenge and exposure frequency \cite{feng2020testing2}. On the aspect of improving the maneuver challenge. Koren et al. \cite{koren2018adaptive} proposed modeling complex interaction scenarios as a Markov decision process, incorporating randomness into the reward function and applying this model to a neural network to capture low-probability events. Karimi et al. \cite{karimi2022automatic} introduced a framework for describing the complexity and modeled the generation of test cases as a constraint satisfaction search problem. On the aspect of enhancing the exposure frequency, Zhao et al. \cite{zhao2016accelerated} proposed an accelerated lane-change scenario testing method based on the importance sampling theorem. Feng et al. proposed a method combining importance sampling with key scenario-intensive reinforcement learning, theoretically proving that this approach can accelerate test efficiency by up to 2000 times under natural traffic flow on public roads \cite{feng2023dense} \cite{liu2024curse}. However, these methods primarily adopt a hybrid approach where the VUT operates in the test track, while other participants are generated through a virtual platform. This approach still faces challenges in modeling the exact and complex interaction scenarios.

Although significant progress has been made in track testing, it faces two primary challenges, which also represent significant barriers. (i) The object targets used in track testing are limited by rigid control methods and low intelligence, often involving dummy targets against intelligent ADS. Additionally, the setup of these tests is costly, labor-intensive, and time-consuming. Existing test equipment mainly employs track-based or pre-set trajectory control methods, which are typically designed for ADAS testing and are not well-suited for high-level ADS testing \cite{fremont2020formal}. (ii) Current testing methods are inefficient and lack interactivity. It is impractical to account for every potential scenario an automated vehicle could face within a fixed set of conditions, as the range of possible interactions is virtually limitless. However, most existing studies overlook the dynamic interactions between automated vehicles and other road users in complex traffic environments \cite{wei2024interactive}. Furthermore, the majority of studies focus primarily on scenario design, with limited research on adversarial testing in real-world scenarios. Consequently, these testing methods fail to meet the growing need for ADS testing in interactive and dynamic environments.

In response to the emerging demands and challenges in ADS testing, this paper proposes a novel testing framework, Real-world Troublemaker, designed for ADS testing on test tracks to address the issues identified earlier. The proposed framework leverages cloud-controlled technology to generate safety-critical test scenarios. Through real-time communication between object targets and the cloud control center, flexible control of multiple object targets is achieved, offering an efficient, automated, and cost-effective solution for ADS track testing. To fulfill the adversarial testing requirements for risk-interactive scenarios, a dynamic interaction game strategy is developed based on a quadratic risk utility function for maneuver challenge. Furthermore, importance sampling is introduced to enhance the exposure frequency of safety-critical scenarios, facilitating the accelerated testing of real interactive scenarios in a closed-course environment.

The contributions are summarized as follows:
\begin{itemize}
    \item Design a novel 5G cloud-controlled track test framework. The framework is capable of generating highly interactive scenarios, enabling remote control of the testing process. It addresses the limitations of existing test tracks, such as labor-intensive setups, time-consuming preparations, and poor dynamic interaction.
    \item Propose an interactive testing scenario generation strategy, which designs a dynamic game model based on quadratic risk interaction utility to enhance maneuver challenges, along with a risk scenario exposure distribution function based on importance sampling. This provides a novel model to address the bottlenecks, such as low efficiency and difficulty in scenario reproduction.
    \item Successfully deploy and validate a Troublemaker testing system, the first cloud-controlled track testing framework for ADS, contributing to the development of track testing for high-level ADS.
\end{itemize}

The rest of this paper is organized as follows. In Section \ref{sec:Overview of the Troublemaker}, the overview of the Troublemaker is presented. Then, an interactive concrete scenario generation method is proposed in Section \ref{sec:interaction concrete scenario generation}. In Section \ref{sec:implementation of the Troublemaker}, the implementation of Troublemaker is presented. The field tests and the results are analyzed in Section \ref{sec:track testing analysis}. Finally, Section \ref{sec:Conclusion and future research} concludes the paper.

\section{Overview of the Troublemaker}\label{sec:Overview of the Troublemaker}
\begin{figure}[!t]
\centering
\includegraphics[width=3.3in]{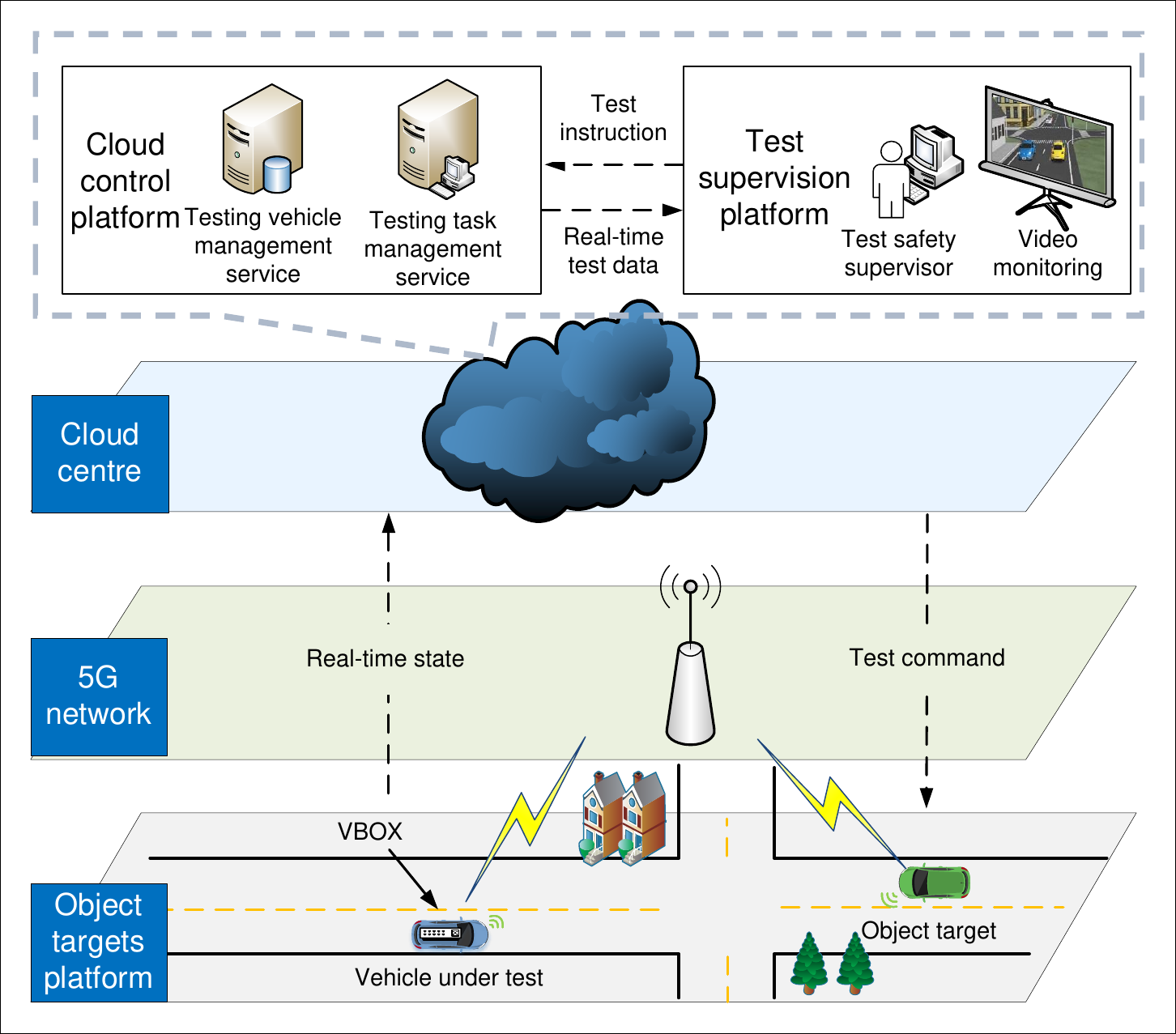}
\caption{Overview of the Troublemaker.}
\label{fig:Overview of the Troublemaker.}
\end{figure}

As shown in Fig. \ref{fig:Overview of the Troublemaker.}, the overall architecture of the Troublemaker is illustrated. This framework consists of three main platforms: a cloud control platform, an object target motion platform, and a test-dedicated 5G network. Additionally, a test monitoring and command center, as well as a VUT test box (VBOX), have been developed for issuing test instructions and providing remote access to test personnel.

The cloud control platform is the core of the entire framework and includes three primary modules: a data transmission protocol encoding/decoding module, a test scenario management module, and a cloud database. The encoding/decoding module handles the data packaging process according to the defined communication interface protocol and the cloud-monitoring center test instruction interface protocol. The test scenario management module is responsible for generating interactive test scenarios. It generates challenging and high-interaction scenarios for the VUT by planning the trajectories of object targets. The specific methodology is described in Section \ref{sec:interaction concrete scenario generation}. Additionally, real-time test data is uploaded to the cloud database for storage and recording. The communication network serves as the medium for system information exchange, using a dedicated 5G network to enable information interaction and data sharing. The object target platform, comprising mimic traffic participants that interact with the VUT, is composed of a motion chassis and a vehicle shell. The motion chassis is capable of executing trajectory commands from the cloud control platform while simultaneously uploading real-time motion state information to the platform. The test monitoring and command center, along with the VBOX vehicle collection device, are accessories designed to facilitate remote monitoring and issuing of test instructions. The VBOX is used to collect real-time state information of the VUT without the need for knowledge of the VUT's algorithm or interface protocol.
 During the testing phase, remote test instructors issue commands to the VUT, which then activates its ADS and proceeds to follow the predefined route. Simultaneously, the cloud control center acquires real-time state information of both the VUT and the object targets, computes the motion trajectory for the object targets, and dispatches control commands to execute their movements. This framework facilitates active conflict and adversarial interaction testing with the VUT. A comprehensive explanation of the test workflow is provided in Section \ref{sec:implementation of the Troublemaker}.
\section{interaction concrete scenario generation}\label{sec:interaction concrete scenario generation}
In this section, we focus on the methods for generating interaction concrete scenarios. According to the ISO 34501 standard, a logical scenario is described with the inclusion of parameters, whose values are defined as ranges. A concrete scenario is depicted with explicit parameter values, describing physical attributes. In accordance with the definition of this standard, this study focuses on analyzing the characteristics of the interaction scenario, designing dynamic interaction strategies, and improving the exposure frequency of risk scenarios. The goal is to enable the generation of interaction concrete scenarios and their on-site execution.

\subsection{Interaction Scenarios Characteristics Analysis}
Interaction risk scenarios occur when traffic participants simultaneously occupy road resources, leading to potential conflicts. In such scenarios, participants must consider both the constraints imposed by their counterpart’s decisions and the potential responses to their own decisions \cite{jia2023interactive}. Additionally, individual participants exhibit heterogeneous driving styles, and collision risks can take various forms. These scenarios are typical in real-world traffic, such as unprotected left turns at intersections and merging/diverging in roundabouts. The conditions for these scenarios are: (1) a potential conflict point exists between the participants, creating a risk scenario; (2) the interaction strategies of both parties increase the likelihood of a collision. Based on these features, a probabilistic model for the occurrence of specific interaction scenarios will be developed.

The purpose of tracking testing is to reproduce risk events in interactive scenarios, which can be used to effectively test VUT. An event of interest with VUTs can be denoted as \textit{E} (e.g., accident event). Our goal is to maximize the probability of conflict events occurring, which can be formulated as:
\begin{equation}
\label{eq:eq1}
\max P(E|\theta )
\end{equation}
\begin{equation}
\label{eq:eq2 event}
P(E|\theta )=\sum_{\xi \in \chi }^{} P(E|\xi ,\theta )P(\xi |\theta )
\end{equation}
where $\xi$  is used to describe dynamic elements during a scenario, while $\theta $ represents the parameters determined by the operational design domain (ODD), e.g., number of lanes, road type, weather conditions, \textit{etc}. As shown in (\ref{eq:eq2 event}), the occurrence of an event is expressed as a combination of maneuver challenges $P(E|\xi ,\theta )$ and exposure frequency $P(\xi |\theta )$. However, in real-world traffic environments, due to the complexity of scenarios and the diversity of conflict types, a naive search for safety-critical scenarios faces the problem of dimensional explosion \cite{liu2024curse}. 

This concrete scenario generation can be considered as designing a behavior policy that enables targets to aggressively interact with the VUT. Therefore, by selecting key interaction scenario parameters, a dynamic game approach can perform well in solving this issue. Intersections, as typical high-risk areas with frequent interactions in urban traffic environments, present even more challenges for automated driving systems, especially in unprotected left-turn scenarios \cite{li2024autonomous}. Hence, we use this scenario as an example to introduce the testing method for interaction scenarios. For the purpose of this study, some assumptions are made.

\textit{Remark 1:} This paper focuses on the interaction scenarios between the VUT and traffic participants in traffic environments, specifically safe-critical scenarios. Therefore, static factors such as weather and road conditions are not considered.

\textit{Remark 2:} The research aims to generate strong interaction scenarios involving multiple object targets, primarily focused on generating concrete test scenarios from logical ones. Thus, the generation methods of logical or other more abstract scenarios are not considered.

\textit{Remark 3:} VUT exhibits generic driving behaviors, sharing common "generic features," such as maintaining safe distances and interacting safely with surrounding vehicles, as well as exhibiting driving regularity and smoothness.

\begin{figure*}[!t]
  \centering
\includegraphics[width=5.8in]{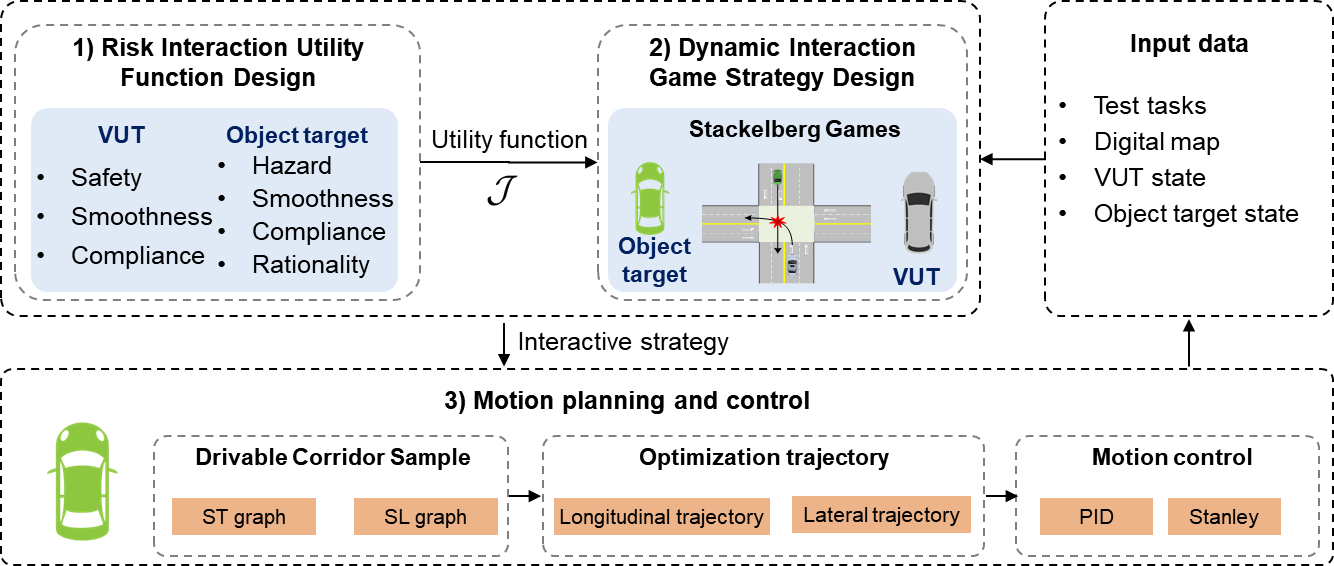}
\caption{Overview of interactive scenario generation based on dynamic game strategy.}
\label{fig:Interactive critical scenario generation based on game strategy.}
\end{figure*}
\subsection{Maneuver Challenge Improvement Through a Dynamic Game Framework}

We propose a game-theoretic model that directly captures interaction dynamics and designs object targets motion control strategies to address enhanced interaction risks. Specifically, this includes the design of a risk interaction utility function, the development of a dynamic interaction strategy based on leader-follower game theory, and object target trajectory planning and control, as shown in Fig. \ref{fig:Interactive critical scenario generation based on game strategy.}.

A dynamic game scenario can be described as \(\mathcal{G} =< \mathcal{P}, \mathcal{X},\mathcal{A},\mathcal{R}>\), where the participants \(\mathcal{P} \)
contains one VUT and object targets, state space \(\mathcal{X} \), \(\mathbf{x} =[s,\dot{s} ,l,\dot{l} ]^{T} \) represents the state of participant, \(\mathbf{x} \in \mathcal{X} \). Our testing scenario is applied to structured road scenarios, where the state is represented by Frenet coordinates \((\mathit{s,l})\). \(\mathit{s}\)  represents the longitudinal distance along the reference path, and \(\mathit{l}\)  represents the lateral deviation along the tangent to the reference path. The evolutions of participants state are described by their trajectories \(\xi _{i}:[0,T] \longrightarrow \mathcal{X} \). Action space \(\mathcal{A} \), \(\mathbf{a}  =[\ddot{s} ,\ddot{l} ]^{T} \), \(\mathbf{a}  \in \mathcal{A} \). Reward function \(\mathcal{R} \), in track testing, a brute-force search is inefficient in specific scenarios and may result in unrealistic phenomena that do not occur in real traffic, such as the object target  directly colliding with the VUT without following traffic rules, where a multi-attribute reward function is designed to address this issue.

To characterize participants state transition, a simplified discrete model is used to represent the vehicle’s kinematics.
\begin{equation}
\label{eq:eq3 Kinem}
\mathbf{\mathit{\textbf{x}} } (k+1)=\mathbf{A} \mathbf{\mathit{\textbf{x}(k)} } +\mathbf{B} \mathbf{\mathit{\textbf{a}(k)} } 
\end{equation}
where,\begin{align}
\mathbf{A} =\begin{bmatrix}
  1  &\Delta t  &0 &0 \\
  0&  1&  0& 0\\
  0&0  &1  &\Delta t \\
  0&0  &0  &1
\end{bmatrix},
\mathbf{B} = \begin{bmatrix}
  \frac{\Delta t^{2} }{2} & 0 \\
  \Delta t & 0 \\
  0 & \frac{\Delta t^{2} }{2}\\
  0 & \Delta t
\end{bmatrix}
\end{align}

\subsubsection{Risk Interaction Utility Function Design}

A multi-attribute utility function is proposed as a core component in generating interaction scenarios. While ensuring the creation of collision scenarios, it also aims to make the behavior of object targets more rational and realistic. The interaction utility function primarily considers three aspects: hazard, smoothness, and compliance. It is formulated as a linear combination of the costs defined above, expressed as:
\begin{equation}
\label{eq:eq5 reward function}
\mathcal{J} =\omega _{h} \mathit{J} _{h} +\omega _{s} \mathit{J} _{s} +\omega _{c} \mathit{J} _{c}
\end{equation}
where \(\mathit{J} _{h}\) is the hazard reward,  \(\mathit{J} _{s}\)is the smoothness reward,  and \(\mathit{J} _{c}\) is the compliance reward. $\omega _{h}$ , $\omega _{s}$ , and $\omega _{c}$ are the weighting coefficients.

\textbf{Hazard:} To maximize vehicle collision risk, maneuver challenge is estimated by post encroachment time (PET), which calculates the time gap between an actor and another actor in a collision point. As discussed in \cite{laureshyn2010evaluation}, PET is one of the most widely used indices of hazard evaluation, and it is defined as
\begin{equation}
PET(A,B,CP)=|t_{B,CP} -t_{A,CP} |
\end{equation}
where A and B are two actors in this interactive scenario, CP is the collision point between them.
As the PET decreases, the hazard risk increases. Therefore, \(\mathit{R} _{h}\) is designed as:
\begin{equation}
\mathit{J} _{h}=PET(A,B,CP)
\end{equation}

However, the PET metric, when used as the sole objective, overlooks the diversity of conflict types and the varying risk levels associated with different scenarios. Real-world interaction environments are characterized by a variety of types and differences in risk severity. Therefore, it is essential to develop a quantifiable multi-type risk interaction utility function for the object targets, addressing the limitations of traditional testing with fixed parameters. To illustrate the method of indicator design, we take the unprotected left-turn scenario as a typical example.
\begin{figure}[!t]
\centering
\includegraphics[width=3.2in]{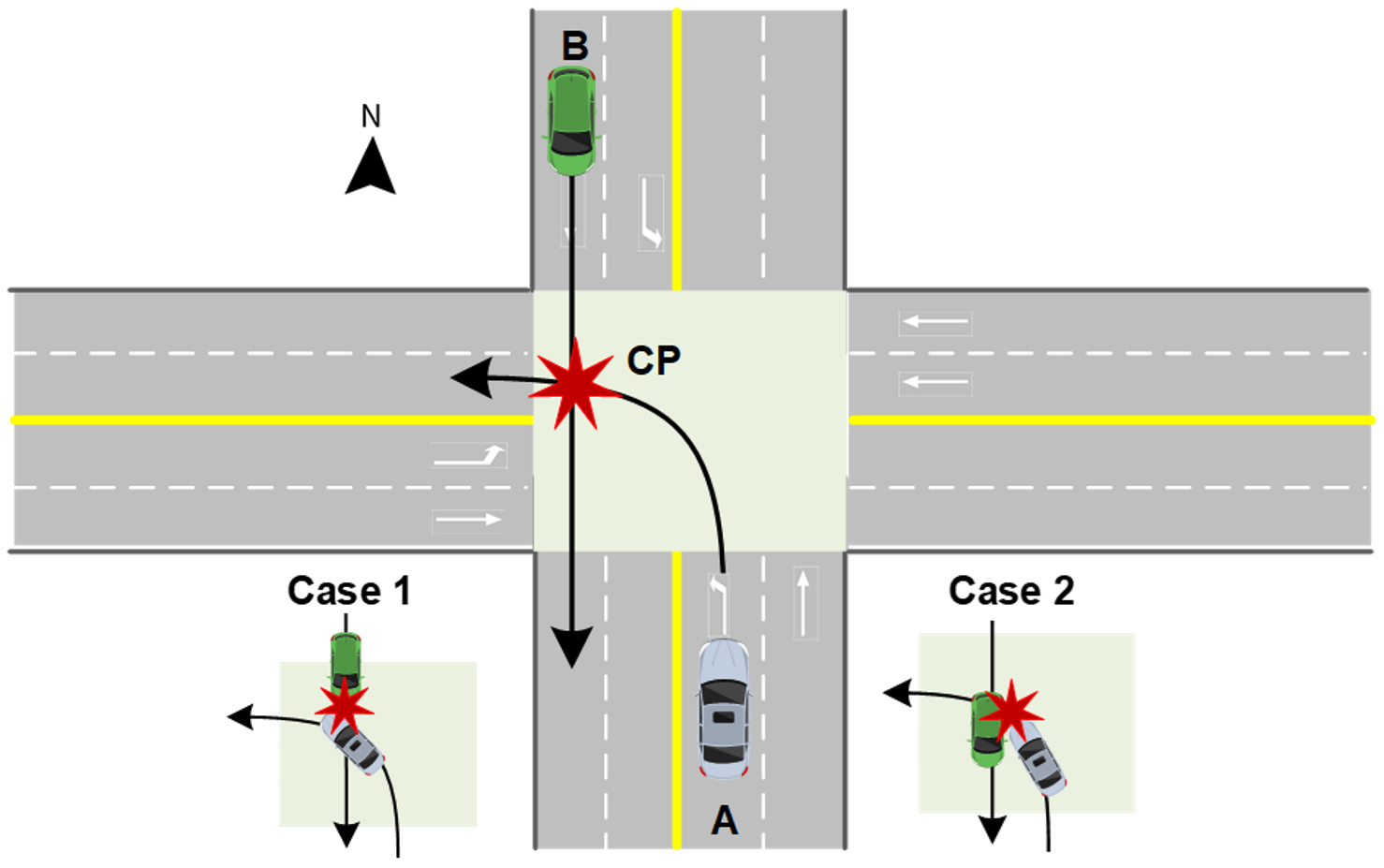}
\caption{The interactive risk illustration.}
\label{fig3}
\end{figure}
As shown in Fig. \ref{fig3}, Vehicle A is moving north in the left-turn lane and expects to make a left turn at an unsignalized intersection. Vehicle B is moving south in the straight lane and expects to continue straight through the intersection. The times at which Vehicles A and B reach the collision point (CP) are denoted as $t_1$ and $t_2$, respectively. There are the following two unsafe scenarios for this interaction:

Case 1: Vehicle A does not yield and arrives at the intersection first, but does not leave enough time for Vehicle B, meaning: \(t_{2} -t_{1} \le X_{1} \).

Case 2: Vehicle A yields, allowing Vehicle B to pass through the intersection first, but the time gap between Vehicle B and Vehicle A is too small, meaning: \(t_{1} -t_{2} \le X_{2} \)

\begin{figure}[!t]
\centering
\includegraphics[width=3.4in]{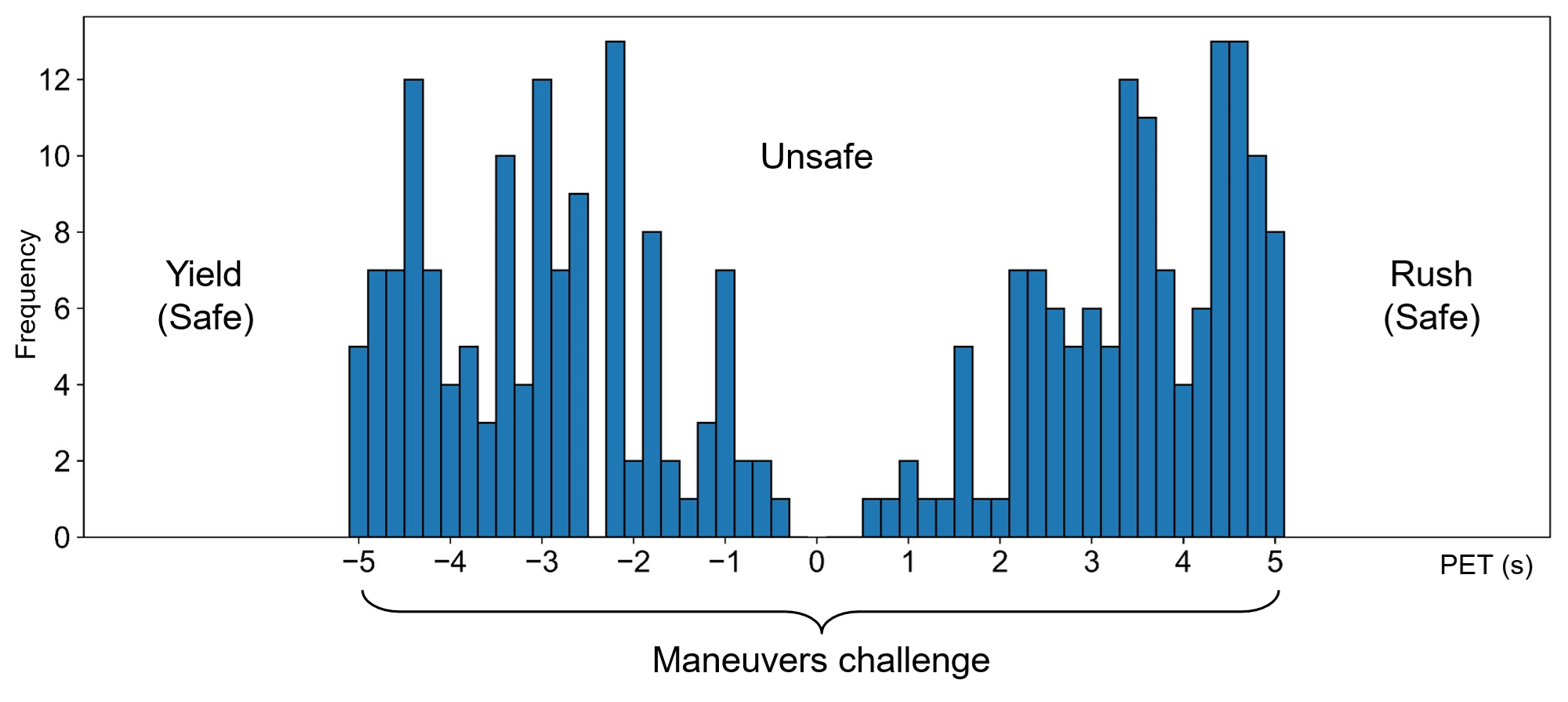}
\caption{Statistical results of empirical driver-vehicle interactions.}
\label{fig:Statistical results of empirical driver-vehicle interactions.}
\end{figure}

Let \(\Delta t=t_{2} -t_{1} \) be defined as the time difference, and use an indicator function \(\mathbb{I} _{PET}(\Delta t) \) to represent the two aforementioned scenarios.
\begin{equation}
\mathbb{I} _{PET}(\Delta t) =\begin{cases}
  1& \text{ if } \Delta t\ge 0 \\
  -1& \text{ if } \Delta t<0
\end{cases}
\end{equation}
where the indicator function is negative, representing a yielding behavior by vehicle A, and positive, indicating that vehicle A is rushing. When the function value falls within the range of the maneuver challenge, it signifies an unsafe collision risk, as shown in Fig. \ref{fig:Statistical results of empirical driver-vehicle interactions.}.

To account for these two scenarios while capturing the varying risk quantification reflecting different human-like characteristics, we propose a quadratic maneuver interaction utility function. It is important to note that this function is designed with the premise of assessing the conflict risk between the object target and the VUT. Through adjustments in driving style, the test scenarios are configured to produce collision risks of varying intensity, simulating the interactive adversarial situations encountered in real-world traffic environments.
\begin{equation}
    \begin{aligned}
    \centering
J_{h}^{\prime} & =h(\psi) p(\psi) \\
& =\frac{1}{m n}\left(\psi^{2}+(m-n) \psi+m n\right) p(\psi)
\end{aligned}
\end{equation}
\begin{equation}
\psi=P E T(\mathrm{~A}, \mathrm{~B}, \mathrm{CP}) \cdot \mathbb{I}_{P E T}(\Delta t)
\end{equation}
where \textit{m} and \textit{n} represent the boundary values for the safe and unsafe regions. $\psi$ 
 represents the PET.

 \textbf{Smoothness:} To ensure the authenticity of the test, the object targets' motion must account for smoothness requirements. In this study, the smoothness metric is quantified as the negative integral of the sum of the squared longitudinal and lateral accelerations over a time period \( T \).

 \begin{equation}
J_{s} =\sum_{k=1}^{T} (\ddot{s}^{2} +\ddot{l}^{2}   )
 \end{equation}

\textbf{ Compliance:} To prevent object targets from displaying unrealistic collision behaviors during the test tasks, a predefined test route is established. The expectation is that the vehicle will follow traffic rules throughout the test.
\begin{equation}
J_{c} =\sum_{k=1}^{T} (\mathbf{x}_{k}-\mathbf{x}_{ref} )^{T} \mathbf{Q} (\mathbf{x}_{k}-\mathbf{x}_{ref})
\end{equation}

\subsubsection{Dynamic Interaction Game Strategy Design Based on Stackelberg Equilibrium}

In a strongly interactive scenario, a dynamic game problem occurs between the VUT and the object target, which is a typical leader-follower game. The VUT is treated as the follower, while the object target  serves as the leader. Through adversarial interaction, both entities make optimal conflict strategies, with each influencing the other's decision-making. This yields:
\begin{equation}
\begin{array}{l}
(\mathbf{u}^{L*}, \sigma^{*})= \underset{\mathbf{u}^{L}, \sigma}{\arg\min} \underset{\mathbf{u}^{F}\in \gamma ^{2}(\mathbf{u}^{L}, \sigma)}{\min} J_{L}(\mathbf{x}^{L}, \mathbf{x}^{F}, \mathbf{u}^{F}, \mathbf{u}^{L}, \sigma) \\
\gamma^2 (\mathbf{u}^{L}, \sigma) \\
= \left\{ \zeta \in \Phi^2 : J_{F}(\mathbf{x}^{L}, \mathbf{x}^{F}, \mathbf{u}^{L}, \sigma, \zeta) \le J_{F}(\mathbf{x}^{L}, \mathbf{x}^{F}, \mathbf{u}^{L}, \sigma, \mathbf{u}^{F}), \right. \\
\quad \quad\left. \forall \mathbf{u}^{F} \in \Phi^2 \right\} \\
\mathrm{s.t.}\quad\sigma \in \left\{ \text{rush}, \text{yield} \right\},  u_{ij} \in [a_{\text{min}}, a_{\text{max}}], v_{ij} \in [0, v_{\text{max}}]
\end{array}
\end{equation}
where $\sigma $ is the strategy of the actors, $\sigma \in \left \{ rush,yield \right \}$, $a_{min}$ and $a_{max}$ are the minimum and maximum acceleration, respectively.   $\textbf{v}$ represents the longitudinal and lateral velocity and $v_{max} $ is the maximum velocity. 

The best response of the follower to the leader’s actions, as obtained from (13), can be solved as an optimal control problem (OCP). However, in our case, the follower's response itself is given as an OCP, resulting in a nested optimization problem. To solve this efficiently, we use model predictive control (MPC) for vehicle state prediction, establishing a time-varying linear system and discretizing it. Furthermore, each agent has its own objective function, denoted by \( J_L \) and \( J_F \). The predictive horizon is \( N \), and the sampling interval is \( \Delta t \), so the state sequence is given by
\begin{equation}
    \begin{array}{l}
\mathbf{x}(k+1)=\mathbf{x}(k)+f(\mathbf{x}(k), \mathbf{u}(k)) \Delta t \\
\mathbf{x}(k+2)=\mathbf{x}(k+1)+f(\mathbf{x}(k+1), \mathbf{u}(k+1)) \Delta t \\
\vdots \\
\mathbf{x}(k+N)=\mathbf{x}(k+N-1)+ f(\mathbf{x}(k+N-1), \\ \mathbf{u}(k+N-1)) \Delta t
\end{array}
\end{equation}
where \( f(*) \) represents the system's state transition equation, and the kinematic model provided in Eq. (1) is used.

For the VUT, the optimization objectives include safety, regularity, and smoothness, while for the object target, the optimization focuses on hazards, regularity, and smoothness, as the goal is to ensure the target vehicle behaves optimally in response to the leader's actions.

Given state sequence $\textbf{x}^F$ of the follower, The cost function $J_F$ is defined by
\begin{equation}
\begin{aligned}
\min_{u_{k:k+N-1}^{F} }  J_F&= J_s+J_c \\
    &= \sum_{i=k}^{k+N-1}(\left \| \mathbf{x}_{i}^{F}-\mathbf{x}_{ref}^{F}\right \|_{\mathbf{Q} }^{2}  ) 
+\left \| \mathbf{u}_{i}^{F}   \right \|_{\mathbf{R} }^{2} \\ 
\text{s.t.} \quad
& \mathbf{x} _{i+1}^{F}=f(\mathbf{x} _{i}^{F},\mathbf{u} _{i}^{F}  ) ,\quad u_{ij}\in [a_{min},a_{max}] \\
&s_{i}^{F} -s_{safe}-s_{i}^{L}  \ge 0\\
&l_{i}^{F} -l_{safe}-l_{i}^{L}  \ge 0\\
 \end{aligned}
\end{equation}

Given state sequence $\textbf{x}^L$ of the leader, The cost function $J_L$ is defined by
\begin{equation}
\begin{aligned}
\min_{u_{k:k+N-1}^{L} } J_L&= J_s+J_c+J_h \\
    & = \sum_{i=k}^{k+N-1}(\left \| \mathbf{x}_{i}^{L}-\mathbf{x}_{ref}^{L}\right \|_{\mathbf{Q} }^{2}  
+\left \| \mathbf{u}_{i}^{L}   \right \|_{\mathbf{R} }^{2} \\ 
&+\frac{1}{mn}(\psi ^{2}+(m-n)\psi +mn )p(\psi ))      \\
\text{s.t.} \quad
& \mathbf{x} _{i+1}^{L}=f(\mathbf{x} _{i}^{L},\mathbf{u} _{i}^{L}  ) ,\quad \mathbf{u} _{i}\in [a_{min},a_{max}] \\
&(\mathbf{x^{F},\mathbf{u}^{F}   } ) =J_{F}^{*}(\mathbf{x}^{L},\mathbf{x^{F},\mathbf{u}^{F}   }   ) 
 \end{aligned}
\end{equation}

Clearly, the follower's objective function is a typical quadratic optimization problem, and the constraints are affine within the feasible domain. Therefore, the follower's optimal control problem (OCP) is a typical convex optimization problem. Henceforth, the bi-level optimization can be reformulated as a single level problem by using Karush-Kuhn-Tucker (KKT) condition: 

\begin{equation}
    \begin{array}{ll}\min _{\textbf{u}^{L}, \textbf{u}^{F}} & J_{L}\left(\textbf{x}^{L}, \textbf{x}^{F}, u^{L}\right) \\\text { s.t. } & \mathbf{h}_{L}\left(\mathbf{x}^{L}, \mathbf{x}^{F}, \mathbf{u}^{L}\right)=0 \\& \mathbf{g}_{L}\left(\mathbf{x}^{L}, \mathbf{x}^{F}, \mathbf{u}^{L}\right) \leq 0 \\& \mathbf{h}_{F}\left(\mathbf{x}^{L}, \mathbf{x}^{F}, \mathbf{u}^{F}\right)=0 \\& \mathbf{g}_{F}\left(\mathbf{x}^{L}, \mathbf{x}^{F}, \mathbf{u}^{F}\right) \leq 0 \\& \nabla_{\left(\mathbf{x}^{F}, \mathbf{u}^{F}\right)} L\left(\mathbf{x}^{L}, \mathbf{x}^{F}, \mathbf{u}^{F}, \lambda, \boldsymbol{\mu}\right)=0 \\& \lambda \geq 0, \boldsymbol{\mu} \geq 0 \\& \lambda \mathbf{h}_{F}\left(\mathbf{x}^{L}, \mathbf{x}^{F}, \mathbf{u}^{F}\right)=0 \\& \boldsymbol{\mu g}_{F}\left(\mathbf{x}^{L}, \mathbf{x}^{F}, \mathbf{u}^{F}\right)=0\end{array}
\end{equation}

To ensure conciseness, we denote the constraints imposed by the vehicle kinematics model as \textbf{h}, and other bound constraints are represented by the inequality \textbf{g}. 

The optimization objectives for VUT and the object target are based on the designed reward function, where VUT aims to optimize safety, rule adherence, and smoothness, while the object target optimizes adversarial, rule adherence, and smoothness. The problem is solved using the Cadasi solver to obtain the optimal strategy for the target vehicle. Furthermore, during a single test, the VUT and the target vehicle do not engage in an unrestricted adversarial game, especially when both VUT and the target vehicle adopt a wait-and-yield strategy, which could lead to a deadlock situation. To avoid invalid adversarial interactions, we introduce an adversarial rationality check function. If the game duration \textit{M} exceeds the maximum interaction threshold $\delta$, the game is terminated. The algorithm for generating Adversarial Game Scenarios is summarized in Algorithm 1.

\begin{algorithm}[!b]
\caption{Algorithm of Adversarial Game Scenarios generating}
\label{algorithm}
\KwIn{digital map, current state of target vehicle $\mathbf{x_0}^{L}$, current state of VUT $\mathbf{x_0}^{F}$ } 

\KwOut{Optimal strategy and state sequence $\mathcal{\pi}^{L^*}
$}
\BlankLine
State space discretization $N\leftarrow (int)T/\tau$\;
Initialization parameters $J_F\leftarrow0$,$J_L\leftarrow0$,$M\leftarrow0$\;
\While{not reached the maximum times and not passed the CP}{
\For{$i \leftarrow 0$ \KwTo $N$}{

$J_F \gets J_F + (\textbf{x}_F{(k+1)} - \textbf{x}_F{ref})^T \textbf{Q} (\textbf{x}_F{(k+1)} - \textbf{x}_F{ref}) + \textbf{u}_F(k)^T \textbf{R} \textbf{u}_F(k)$\;
$J_L \gets J_L + (\textbf{x}_L{(k+1)} -\textbf{ x}_L{ref})^T \textbf{Q} (\textbf{x}_L{(k+1)} - \textbf{x}_L{ref}) + \textbf{u}_L(k)^T \textbf{R} \textbf{u}_L(k)$\;
$J_h\leftarrow h(\psi )p(\psi )$\;
}

$J_L\gets J_s+J_c+J_h$\;
$Lagrangian(\textbf{x})\gets J_F(\textbf{x})+\lambda\cdot \textbf{h}_F(\textbf{x})+\mu \cdot \textbf{g}_F(\textbf{x})$\;
$\nabla_{\mathbf{\textbf{x}}} L(\mathbf{x}, \lambda, \mu) \gets 0$ \;
$M\gets M+1$\;
}
\end{algorithm}

\subsubsection{Motion Planning and Control}

The motion planning and control problem is decoupled into lateral and longitudinal OCP along the reference path. Based on the upper-level game results, the optimal state sequence for the target vehicle in the future decision time domain is obtained, and the decision result is represented as a sequence of state points in the frenet coordinate system. To obtain the actual trajectory of the target vehicle and complete the tracking control, a sample based optimization method is used to find an optimal trajectory that satisfies the motion constraints for both lateral and longitudinal directions. The optimal driving path sequence and speed sequence are derived separately, and the trajectories are combined to form the actual driving trajectory. Taking the future state points from the game result as the target, a space sampling method is performed between two adjacent state points. SL graph and ST graph are designed to form the lateral and longitudinal drivable corridors, respectively. The surrounding obstacles (VUT, other targets) are projected onto the graphs, and the reference trajectory that satisfies the constraints and avoids collisions is selected based on the evaluation function. After solving the lateral and longitudinal planning problems, the optimal state sequences are obtained, and a safe, comfortable, and stable optimal motion sequence is synthesized in the frenet coordinate system as the final output trajectory. This trajectory is then transmitted to the target vehicle’s control system via the vehicle-cloud link, where longitudinal control is performed using PID, and lateral control is achieved using Stanley for tracking.

\subsection{Risk Scenario Exposure Enhancement Based on Importance Sampling}
In actual driving, interactive behaviors are heterogeneous, and conflict types are diverse. Therefore, it is necessary to consider the interaction characteristics of traffic participants in real traffic scenarios when designing the risk features of target vehicle interactions. Based on the empirical driver-vehicle interaction data and statistical real-world data provided by the 43267 standard, we plot the frequency histogram of the interaction scene PET, as shown in Fig. \ref{fig:Statistical results of empirical driver-vehicle interactions.}. In the unprotected left-turn scenario, high-risk situations, where PET approaches 0, are rare. Many risk scenarios exhibit a bimodal distribution in yielding and overtaking events. Therefore, to better fit the characteristics of this bimodal distribution, we employ the kernel density estimation method to fit this data. The probability density function established by this method is:
\begin{equation}
    \hat{p} (\psi )=\frac{1}{nh} \sum_{i=1}^{n} K(\frac{\psi -\psi _{i} }{h} )
\end{equation}
where \textit{n} is the number of samples, \( h \) is the bandwidth, and \( K \) is the kernel function. In this paper, a Gaussian kernel function is chosen.

The optimal bandwidth $\mathit{h^{*} } $ is determined by minimizing the mean squared integration error using Silverman's rule \cite{chung2018electric}. The resulting optimal bandwidth is given by:
\begin{equation}
    \begin{matrix}
 \mathit{h^{*} }=0.9\cdot min(\hat{\sigma },\frac{IQR}{1.34}  )\cdot n^{-\frac{1}{5} }  \\
IQR=Q_{3}-Q_{1}  
\end{matrix}
\end{equation}
where \( h^* \) is the optimal bandwidth, \( \sigma \) is the standard deviation of the data, and IQR is the interquartile range, with \( Q_3 \) and \( Q_1 \) representing the third and first quartiles, respectively.
\begin{figure*}[!t]
  \centering
\includegraphics[width=5in]{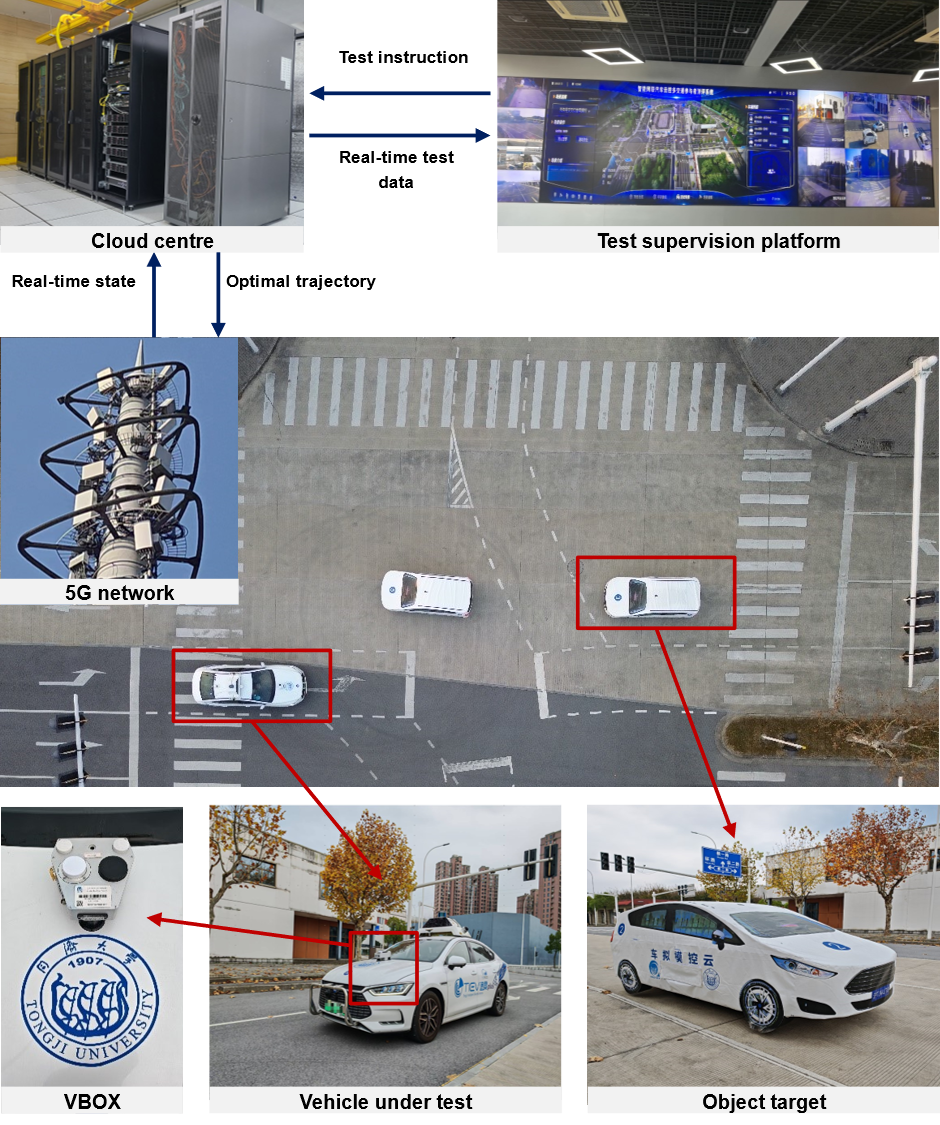}
\caption{The implementation of the Troublemaker.}
\label{fig:The implementation of the Troublemaker.}
\end{figure*}

The exposure frequency of safety-critical scenarios is relatively low, but these scenarios contribute more significantly to the occurrence of collision events. Therefore, we aim to increase the probability of the event of interest occurring. To achieve this, we employ the importance sampling theorem to increase the exposure rate of high-risk scenarios, thereby accelerating the testing process.
\begin{equation}
    \begin{aligned}
E\left(R_{h}^{\prime}\right) & =\int h(\psi) p(\psi) \\
& =\int \frac{h(\psi) p(\psi)}{q(\psi)} q(\psi) \\
& \approx \frac{1}{n} \sum_{i=1}^{n} \frac{h(\psi) p(\psi)}{q(\psi)}, \psi_{i} \sim q(\psi)
\end{aligned}
\end{equation}
where \textit{n} denotes the total number of tests,  $\psi _{i} \sim q(\psi )$ denotes the sampled scenarios according to the distribution.

\section{implementation of the Troublemaker}\label{sec:implementation of the Troublemaker}
The proposed framework has been implemented at the Intelligent Connected Vehicle Evaluation Base of Tongji University. As illustrated in Fig. \ref{fig:The implementation of the Troublemaker.}, the simulated urban block demonstrates the comprehensive implementation of the Real-world Troublemaker system, which integrates multiple components: a testing supervision platform, cloud control centre, 5G network infrastructure, object targets, and VBOX. The testing supervision platform, strategically located at the command center of the testing area, serves as the central hub where test personnel initiate and monitor remote test instructions. The cloud control platform plays a pivotal role in generating dynamic test scenarios, orchestrating the motion of object targets, and managing scenario resets in response to received instructions. Following computational processing, the optimal trajectory for object targets is transmitted for execution. Concurrently, critical data streams including VBOX (vehicle state data) and object target information are continuously uploaded to the cloud control center in real-time. The testing supervision platform maintains comprehensive access to both real-time video streams and test data from the field, facilitated by a robust 5G and fiber-optic network infrastructure. The communication architecture employs the message queuing telemetry transport (MQTT) protocol as the middleware for data transmission between targets and the cloud. The cloud system actively monitors specific topics to acquire vehicle state data, which is then processed through third-party algorithms to generate vehicle trajectories. These trajectories are subsequently transmitted back to the vehicles via the MQTT protocol. The dedicated 5G communication network, operating within the 949.5–959.5 MHz frequency band, has been specifically deployed for the test field, with 5G base stations strategically positioned throughout the testing area. The cloud-controlled target vehicle is equipped with an advanced, self-developed cloud-controlled chassis system, incorporating three core components: a high-precision positioning module, a 5G customer premises equipment (CPE), and a sophisticated vehicle control unit. The object targets maintain continuous communication with the cloud center, transmitting real-time position and orientation data at a frequency of 20 Hz through the 5G CPE. These targets precisely follow the cloud-generated trajectories, with control commands being executed via the CAN bus interface to the vehicle chassis.

\section{track testing analysis}\label{sec:track testing analysis}
\subsection{Experiment Design}
In this section, the experimental design for performance evaluation is presented in detail, including the design of test scenarios, the comparison test method, evaluation metrics, and the configuration of test parameters.

\subsubsection{Design of Test Scenarios} 
\begin{figure}[!t]
\centering
\includegraphics[width=3.4in]{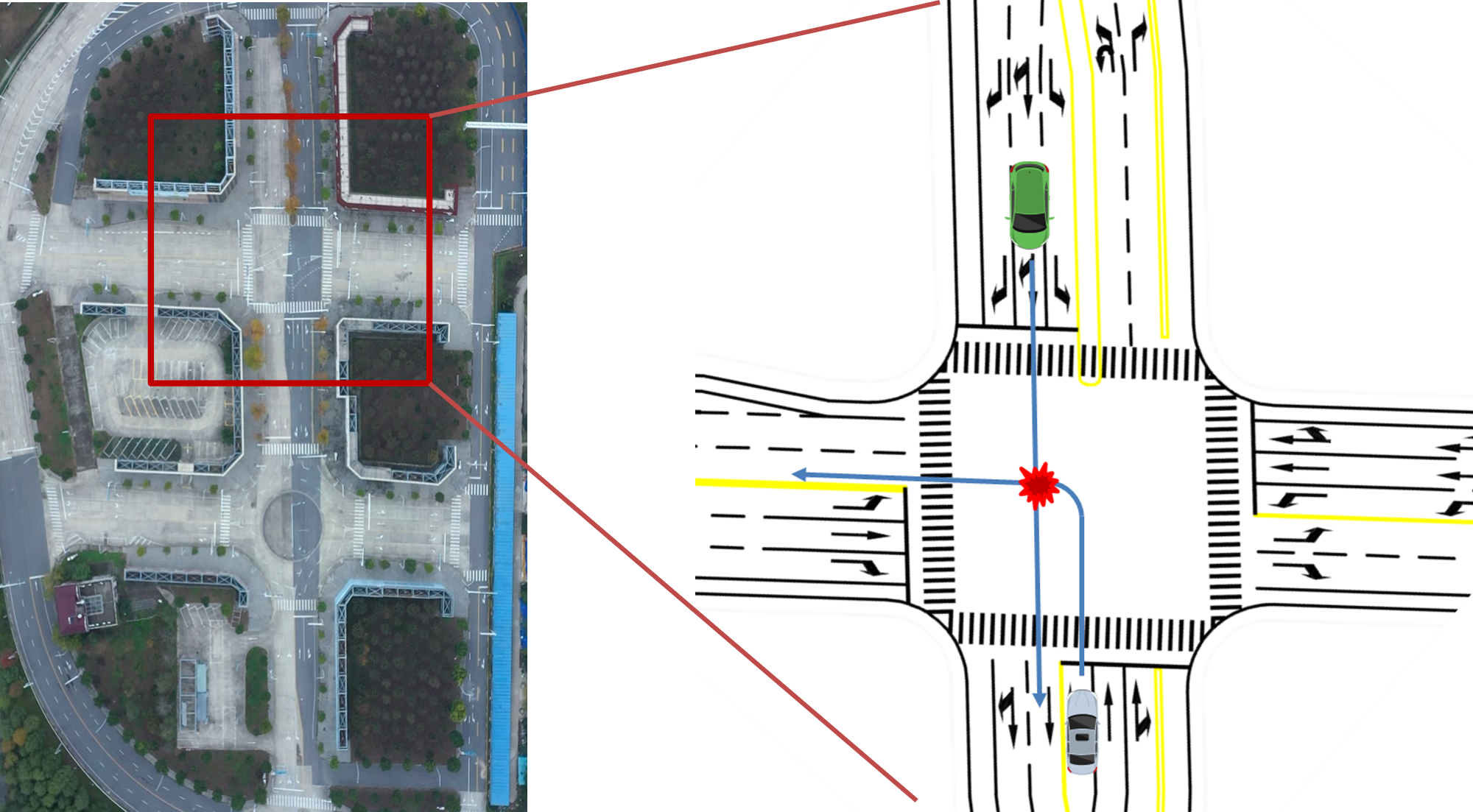}
\caption{Real-world test site scenario design.}
\label{fig:fig6 real-world test site}
\end{figure}

The test site is located at the Tongji University Intelligent Connected Vehicle Evaluation Base, which provides an urban road intersection. The intersection is a cross-shaped, bidirectional 5-lane intersection with a dedicated left-turn lane. An unprotected left-turn scenario is adopted to evaluate the performance of Troublemaker (as shown in Fig. \ref{fig:fig6 real-world test site}). In this scenario, the VUT makes a left turn through the intersection while the object target continues straight through. To maintain its desired velocity, the VUT must interact with the opposing straight-moving object target, with the red dots in the figure indicating the pre-collision points of both vehicles. To validate the effectiveness of the Troublemaker test, field tests will be conducted to assess the accuracy of reproducing real-world scenarios, the diversity of interaction strategies, and the effectiveness of increasing exposure to high-risk scenarios.

\subsubsection{Test Method for Comparison} The predefined baseline test method, which is the typical testing approach currently used at the test site, serves as the comparison method. Relevant parameter values are set based on the site conditions and are described as follows. The VUT drives within its lane and makes a left turn through the intersection. When the pre-collision time between the two vehicles reaches an interval of 4.5 to 5.5 seconds, the object target continues to move straight through the intersection at a constant speed of 3 m/s.

\subsubsection{Evaluation Metrics} The following metrics are utilized for performance evaluation:
\begin{itemize}
\item{\textit{Mean Error of Minimum PET:}} Under multiple tests with the same operating conditions, the mean error between the preset minimum PET for the interaction event and the actual minimum PET observed in testing. This metric indicates the deviation between the actual test results and the expected test outcomes. A smaller value signifies higher test accuracy
\item{\textit{Hausdorff Distance of Object Target Velocity:}} The Hausdorff distance between the object targets' velocity curves across multiple trials has been widely used in time-series trajectory data to measure the similarity and diversity of different sets of time-series data. In this scenario, since the object targets' behavior strategy is primarily reflected in their driving velocity, this metric is employed to characterize the diversity of interaction behaviors \cite{zhou2024evaluating}. A larger Hausdorff distance indicates greater diversity in the interaction behaviors generated across multiple trials. The specific calculation is provided in (\ref{eq:eq21 hausdorff cul}).
\begin{equation}
\label{eq:eq21 hausdorff cul}
    \begin{array}{l}
         H(A,B) =max(h_{d}(A,B),h_{d}(B,A)  ) \\  
         h_{d}(A,B) =max_{a_{i}\in A }(min_{b_{j}\in B}\left \| a_{i}-b_{j}   \right \|) \\
         h_{d}(B,A) =max_{b_{j}\in B }(min_{a_{i}\in A}\left \| b_{j}-a_{i}   \right \|) \\
    \end{array}
\end{equation}
\item{\textit{Incident Rate of Severe Conflicts:}} The proportion of severe conflict events occurring in actual interaction scenarios across multiple tests. In this study, a risk value greater than 0.85 ($PET<2.3$) is defined as a severe conflict. This metric characterizes the challenge of generating adversarial scenarios. A higher incidence rate of severe conflicts indicates a greater exposure to high-risk challenges.
\end{itemize}

\subsubsection{Test Parameter Description} In this experiment, an automated electric vehicle is used as the VUT. The vehicle is equipped with autonomous perception, decision-making, and control capabilities. The object target is a cloud-controlled vehicle, also developed internally. The main parameters of the VUT and object targets are shown in Table \ref{tb:tab1}.

\begin{table}[!t]
\begin{center}
\caption{Parameter values for the VUT and object target}
\label{tb:tab1}
\begin{tabular}{| c | c | c |}
\hline
Parameters & Values\\
\hline
length of object targets & 3.65 m\\
\hline
width of object targets & 1.55 m\\
\hline
target velocity & 3 m/s\\
\hline
length of VUT & 4.76 m\\
\hline
width of VUT & 1.85 m\\
\hline
\textbf{Q} & $diag(1,20,1,5)$\\
\hline
$\omega _{h}$  & 0.7\\
\hline
$\omega _{s}$  & 0.2\\
\hline
$\omega _{c}$  & 0.1\\
\hline
$\delta$  & 10 s\\
\hline
\end{tabular}
\end{center}
\end{table}

\subsection{Result and Discussion}
The results of the track testing evaluation are presented in this section. To minimize the impact of randomness during the testing process, each method was tested multiple times, with no fewer than 30 valid tests. The trajectory, velocity, and real-time PET data of both the VUT and object target were recorded for each test to calculate the evaluation metrics.

The test results confirm that the Troublemaker can achieve the aforementioned objectives: accurately reproduce interaction scenarios, enrich scenario diversity by providing flexible and diverse interaction strategies, and enhance exposure to high-risk scenarios. As shown in the results, our method improves scenario reproduction accuracy by 65.2 \% compared to the baseline, increases the diversity of the object target interactive strategies by approximately 9.2 times, and enhances the exposure to high-risk scenarios by 3.5 times compared to the exposure to medium-risk scenarios in the standard.

Table \ref{tb:tab2} summarizes the average minimum PET error for specific scenarios generated by the Troublemaker and the baseline. During the actual testing conducted, the baseline yielded an average minimum PET error of 2.1 seconds, whereas the Troublemaker generated an average minimum PET error of 0.73 seconds. The Troublemaker has demonstrated remarkable accuracy in reproducing safety-critical scenarios and effectively generating anticipated adversarial risk scenarios.

\begin{figure}[!t]
\centering
\captionsetup[subfigure]{font={footnotesize}, labelfont={}} 
\subfloat[]{\includegraphics[width=1.7in]{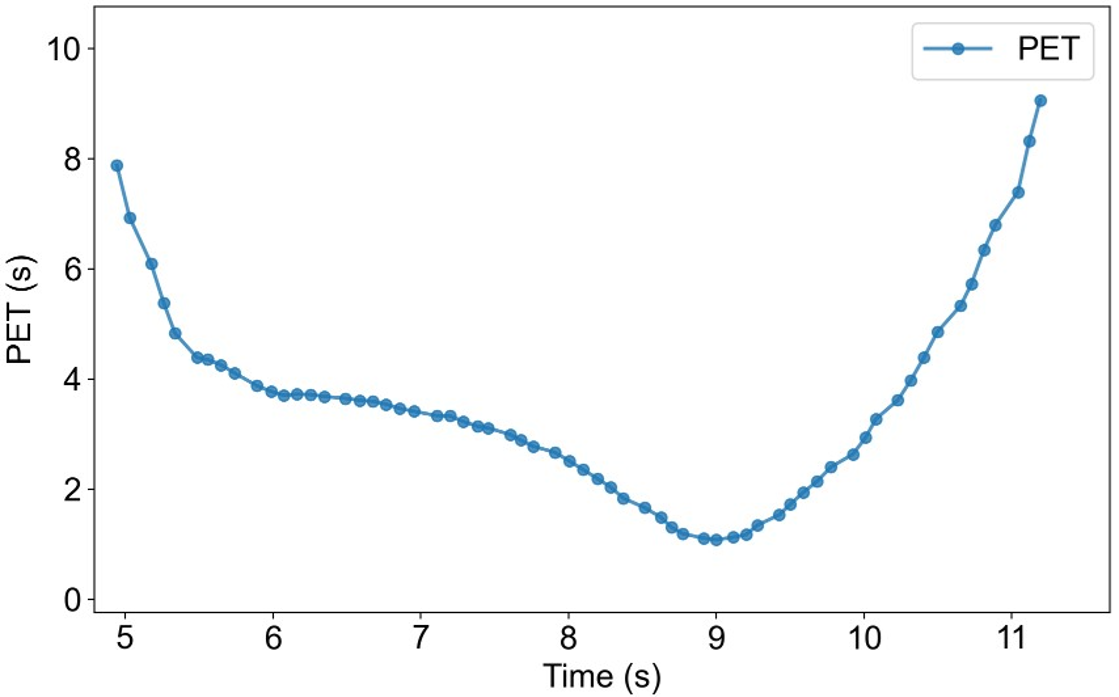}\label{fig:pet_a}}
\hfil
\subfloat[]{\includegraphics[width=1.7in]{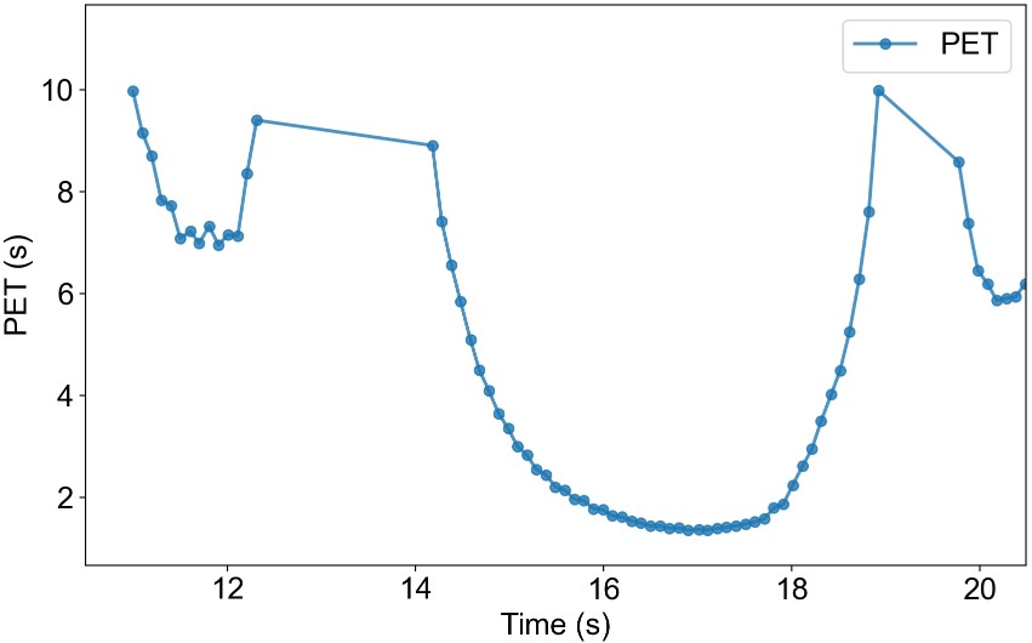}\label{fig:pet_b}}
\caption{The PET results comparing the baseline and the Troublemaker. (a) Baseline. (b) Troublemaker.}
\label{fig:pet comparison}
\end{figure}

Additionally, Fig.\ref{fig:pet comparison} illustrates the real-time changes in PET for typical interactive scenarios generated by both the baseline and the Troublemaker. In the baseline method, the VUT detects the risk and applies braking to avoid the collision, causing the PET to decrease initially and then increase, with a rapid rise after reaching the minimum value. In contrast, in the Troublemaker scenario, the object target dynamically adjusts its strategy, leading to a prolonged period of proactive interaction and conflict with the VUT, which more closely reflects real-world traffic dynamics.

\begin{figure}[!t]
\centering
\captionsetup[subfigure]{font={footnotesize}, labelfont={}} 
\subfloat[]{\includegraphics[width=1.7in]{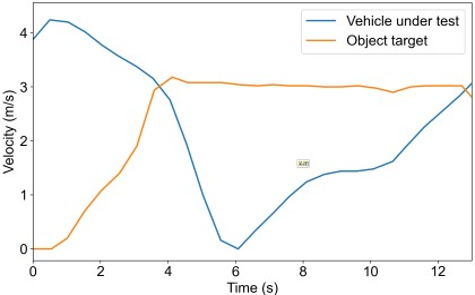}\label{fig:figv_a}}
\hfil
\subfloat[]{\includegraphics[width=1.7in]{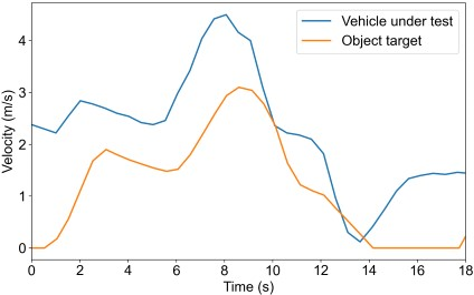}\label{fig:figv_b}}
\caption{The velocity results comparing the baseline and the Troublemaker. (a) Baseline. (b) Troublemaker.}
\label{fig:v comparison}
\end{figure}

Although both the baseline and Troublemaker methods can generate interactive risk scenarios for testing, the baseline method tends to create simple and repetitive scenarios. As shown in the velocity results in Fig. \ref{fig:figv_a}, the baseline method generates risk conflicts with the VUT through a preset approach with fixed parameters, which may lead to "pseudo-interaction" phenomena. In these cases, the ADS avoids potential conflicts using predefined behaviors, thereby preventing actual adversarial interactions or conflicts from occurring. Such behavior is likely based on standard reaction strategies in the test environment, rather than adapting to real-time decisions involving the actions of other traffic participants in complex real-world traffic scenarios. As shown in Fig. \ref{fig:figv_b}, the advantage of the Troublemaker method lies in its ability to generate genuine interactions through proactive behaviors.


\begin{table}[!t]
\begin{center}
\caption{Unprotected left turn scenario test results}
\label{tb:tab2}
\renewcommand{\arraystretch}{1.5}  
\begin{tabular}{|>{\centering\arraybackslash}p{4cm} | c | c |}  
\hline
Metric & Baseline & Troublemaker \\
\hline
Mean error of minimum PET (s) & 2.1 & 0.73 \\
\hline
The maximum jerk value of the VUT ($m/s^3$)  & 1.51 & 2.79 \\
\hline
The velocity hausdorff distance of object targets  & 0.07 & 0.72 \\
\hline
\end{tabular}
\end{center}
\end{table}

In respect of the diversity of the concrete scenarios, the results of the object target's velocity Hausdorff distance (Table \ref{tb:tab2}) support this conclusion. Under the same risk conditions, the Troublemaker produces significantly more diverse behaviors than the baseline. Specifically, the Hausdorff distance for the velocity in the Troublemaker scenario is 0.72, approximately 10 times that of the baseline's 0.07. The baseline method, based on static rules and predefined conditions, generates fixed risk scenario parameters, leading to limited and repetitive interactions among object targets. In contrast, the Troublemaker fosters greater diversity and variation in test scenarios, all while maintaining measurable testing difficulty.

To further demonstrate the actual interaction performance during the adversarial interaction process, two typical scenarios from the track test are selected:
\begin{itemize}
    \item Scenario A: The VUT yields, allowing the object target to pass the intersection first. However, the time gap between vehicle B passing the conflict point is too short, with h = 0.8.
    \item Scenario B: The VUT does not yield and arrives at the intersection first but does not leave enough time for the object target, with h = 0.3.
\end{itemize}

\begin{figure}[!t]
\centering
\includegraphics[width=3.2in]{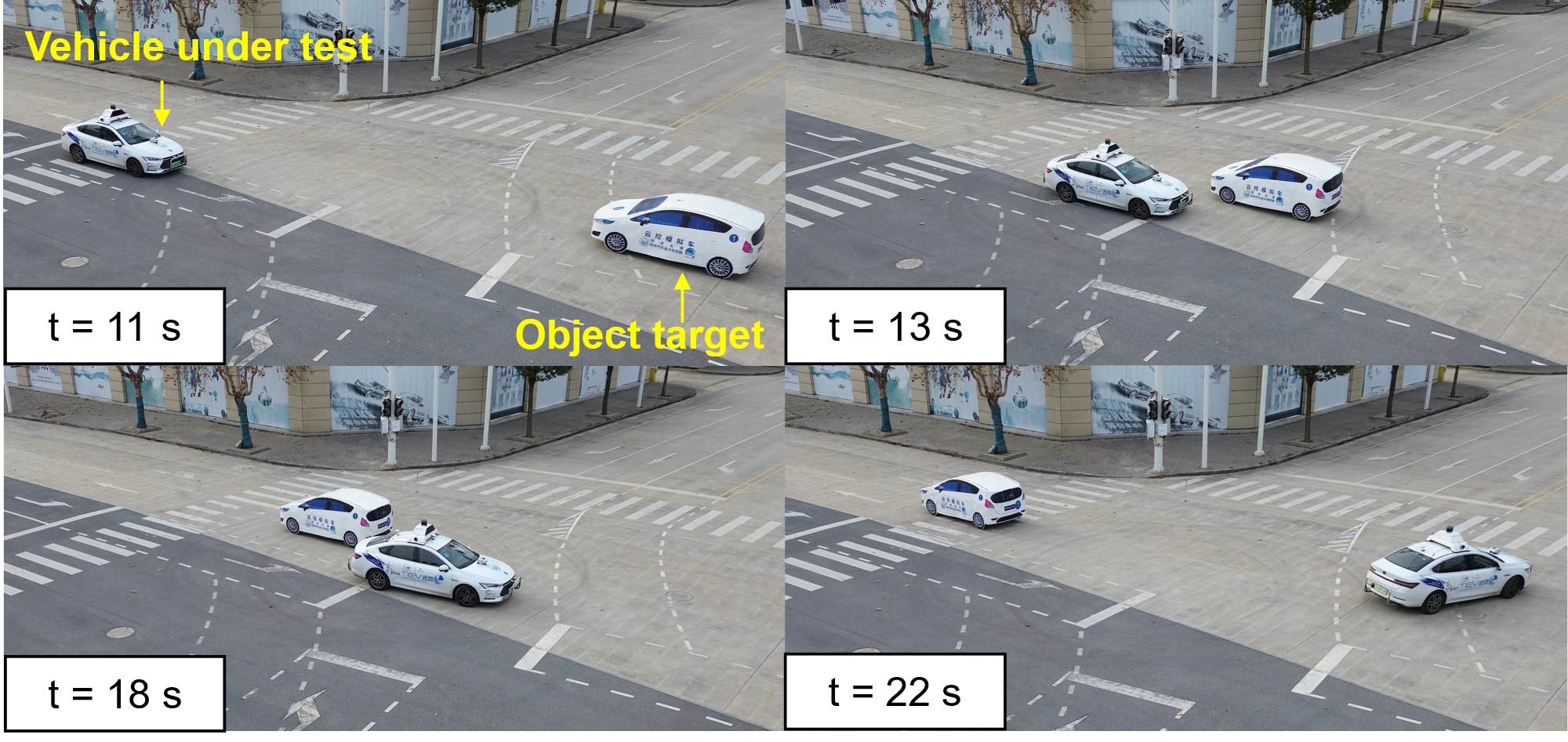}
\caption{The real vehicle test process of scenario A.}
\label{fig:figscenarioAreal}
\end{figure}

\begin{figure}[!t]
\centering
\includegraphics[width=2.6in]{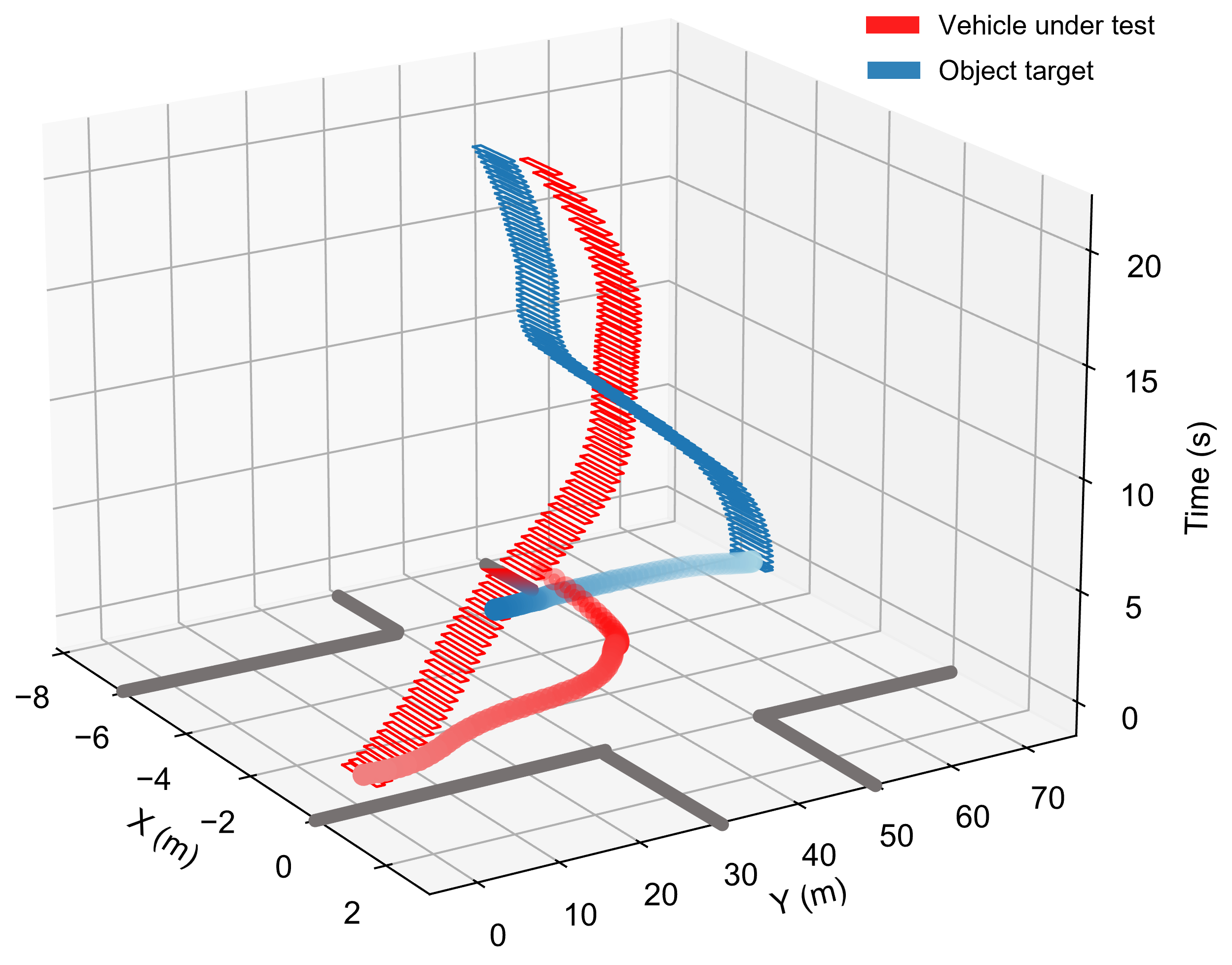}
\caption{Travel trajectories of VUT and object target in scenario A.}
\label{fig:figscenarioA3D}
\end{figure}

Fig. \ref{fig:figscenarioAreal} illustrates the interaction process in Scenario A, spanning 11 seconds. This includes the initial state, the interaction process, and the end of the scenario. During this time, the object target adjusts its speed, forcing the VUT to decelerate and come to a stop to avoid a collision. In this process, the VUT applies braking measures, while the object target dynamically adjusts its speed, creating a sustained adversarial interaction. This continuous interaction prevents pseudo-interaction and only ceases once a reasonable adversarial engagement ends, after which the vehicles exit the intersection. Fig. \ref{fig:figscenarioA3D} presents the actual interaction trajectories, where the \textit{x} and \textit{y} axes represent relative coordinates based on the VUT's starting point in UTM coordinates, and the \textit{z}-axis denotes time. In the potential risk zone, the \textit{z}-axis coordinate of the VUT is higher, indicating that the VUT yields and allows the object target to pass through the conflict zone first.

\begin{figure}[!t]
\centering
\includegraphics[width=3.2in]{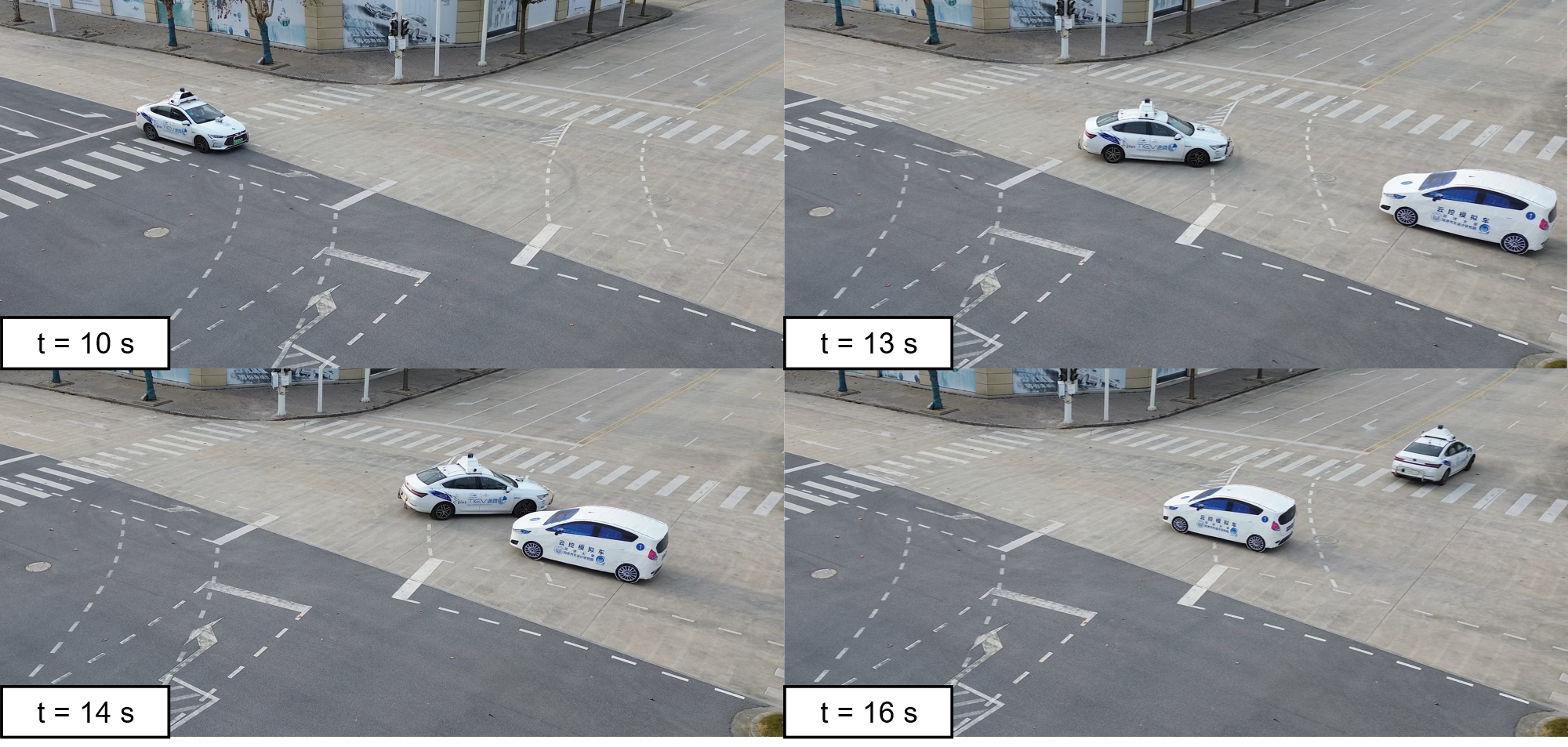}
\caption{The real vehicle test process of scenario B.}
\label{fig:figscenarioBreal}
\end{figure}

\begin{figure}[!t]
\centering
\includegraphics[width=2.4in]{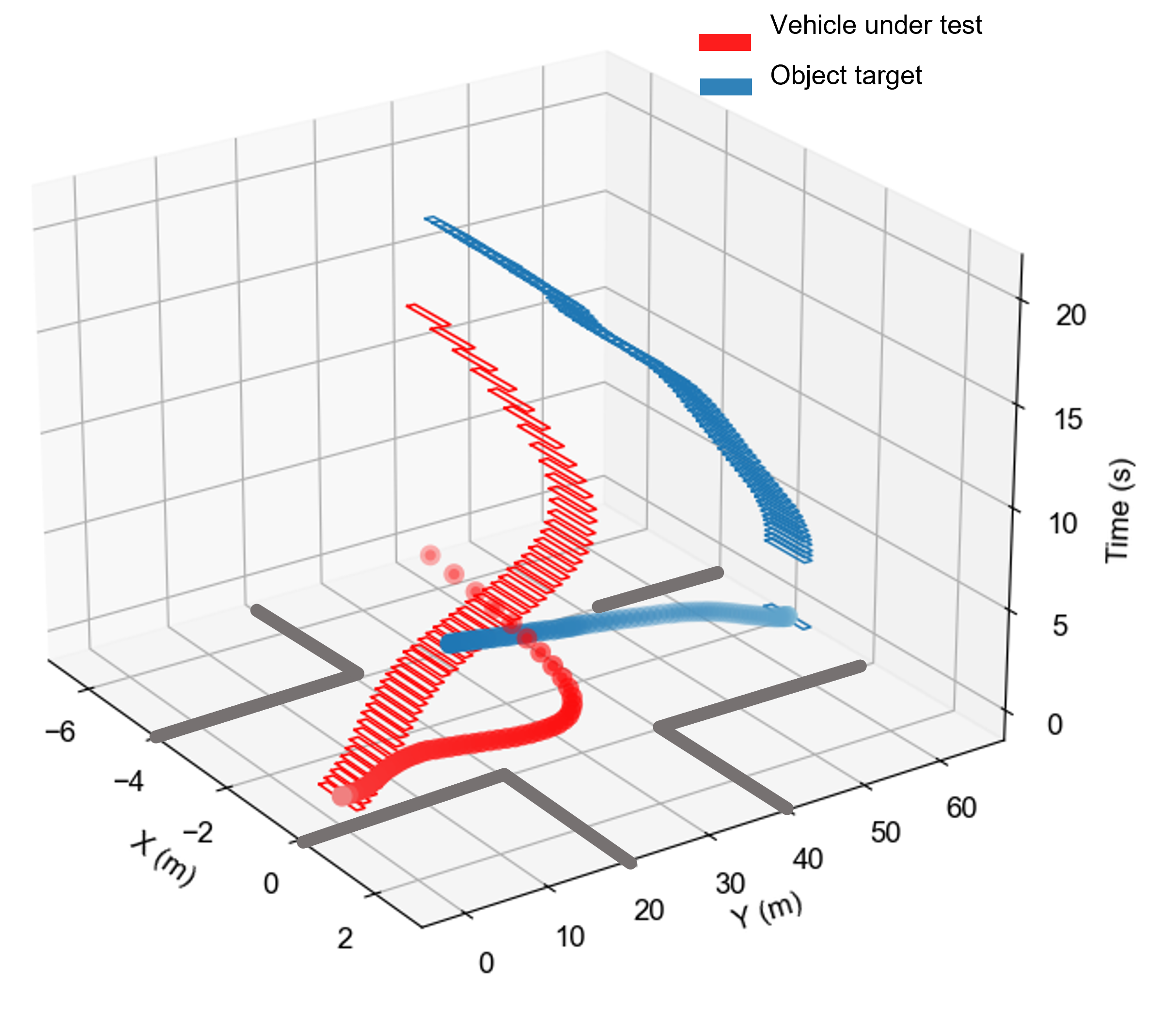}
\caption{Travel trajectories of VUT and object target in scenario B.}
\label{fig:figscenarioB3D}
\end{figure}

Fig. \ref{fig:figscenarioBreal} shows the interaction process in Scenario B, which lasts 6 seconds. In this case, the object target yields, and the VUT determines it can pass through the intersection without stopping, adopting a rush strategy. Fig. \ref{fig:figscenarioB3D} provides the actual interaction trajectory for this scenario, where the VUT leads and passes through the conflict zone first.

\begin{figure}[!t]
    \centering
        \centering
        \subfloat[]{\includegraphics[width=0.48\linewidth]{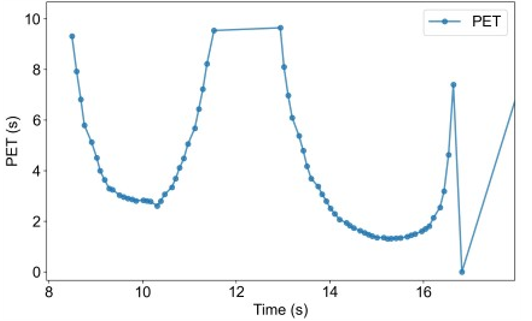}\label{fig:scenario_comparison_a}}
\hfil
        \centering
        \subfloat[]{\includegraphics[width=0.48\linewidth]{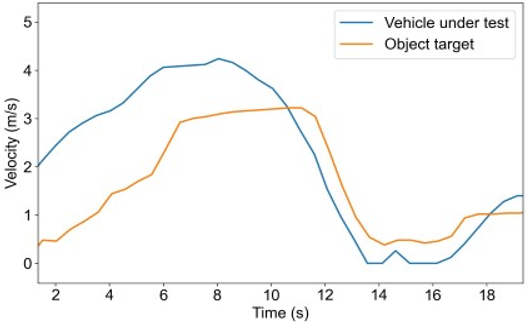}\label{fig:scenario_comparison_b}}
    \vspace{0.01\textwidth} 
        \centering
        \subfloat[]{\includegraphics[width=0.48\linewidth]{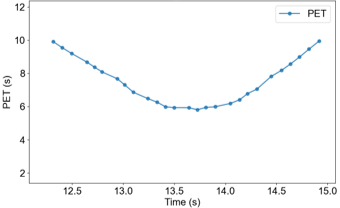}\label{fig:scenario_comparison_c}}
\hfil
        \centering
        \subfloat[]{\includegraphics[width=0.48\linewidth]{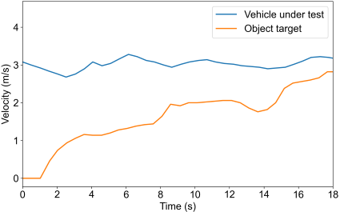}\label{fig:scenario_comparison_d}}

    \caption{Test results in the selected scenarios. (a) The PET results in the scenario A. (b) The velocity results in the scenario A. (c) The PET results in the scenario B. (d) The velocity results in the scenario B.}
    \label{fig:scenario_comparison}
\end{figure}

Fig. \ref{fig:scenario_comparison} illustrates the PET and velocity variation curves for scenarios A and B. As shown in Fig. \ref{fig:scenario_comparison_a} and \ref{fig:scenario_comparison_b}, scenario A, represents a high-risk situation, leading to an adversarial interaction lasting longer than 2 seconds. In contrast, scenario B results in a larger minimum PET. The test results from both scenarios align with the risk quantification utility function and the PET metric, demonstrating consistency in the design. Additionally, Fig. \ref{fig:scenario_comparison_c} and \ref{fig:scenario_comparison_d} present the changes in vehicle speed, revealing that the Troublemaker generates diverse interaction strategies and dynamic speed variations, thereby fostering various types of dynamic, intelligent adversarial interactions with the VUT. An interesting observation from scenario B is that, although no collision occurs between the VUT and the object target during their interaction, the VUT fails to provide a sufficient time gap. Consequently, the object target’s velocity is slightly impacted, prompting it to apply brief braking measures to avoid a collision. This test result provides valuable insight for autonomous vehicle design, suggesting that the harmony of interaction behaviors should be considered. If, in such a scenario, the object target has a long reaction time or fails to brake in time, the VUT could be held responsible for the incident, according to the RSS (Responsibility-Sensitive Safety) framework.

In terms of the high-risk scenario exposure rate, it is observed that severe conflict scenarios are low-probability events in real-world interaction scenarios. To verify the effectiveness of the troublemaker method in high-risk adversarial scenarios, we design a normal distribution as the proposal function. Combined with the risk scenario exposure rate enhancement strategy from Section \ref{sec:interaction concrete scenario generation}, we accelerate the troublemaker test by increasing the exposure frequency of high-risk scenarios and achieve accelerated testing. Similarly, different proposal functions can be selected based on the scenario requirements.

\begin{figure}[!t]
\centering
\includegraphics[width=3.1in]{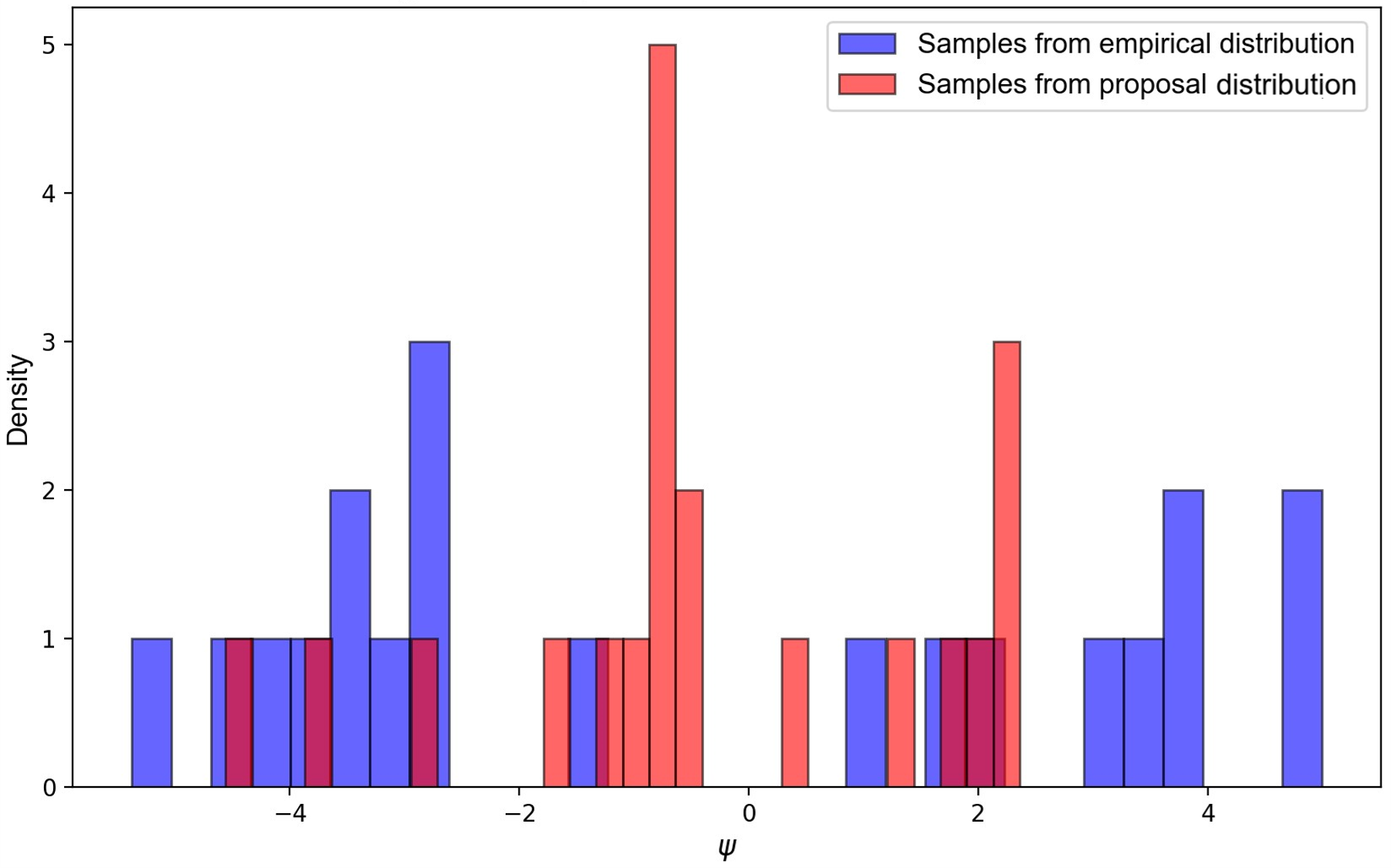}
\caption{Risk distribution of randomly generated test samples.}
\label{fig:risk distribution}
\end{figure}

Fig. \ref{fig:risk distribution} shows that we randomly generated 20 sets of test conditions based on both the empirical distribution and the proposal distribution, and conducted real-world testing based on these conditions. The results indicate that troublemaker can generate more severe conflicts (0.9), which is 3.5 times higher than the real-world data distribution (0.2). Another interesting phenomenon observed in the tests is that as the occurrence of high-risk scenarios increases, we find that despite the object target adopting different decision strategies (yielding or trying to overtake), due to the small PET with the object target and the close actual distance between the two vehicles, the VUT tends to take a braking and evasion strategy more often, which aligns with common sense.

To summarize, troublemaker actively generates dynamic interactions, accurately produces risk interaction scenarios, and effectively ensure the consistency of closed-course test results. Additionally, it can generate diverse interaction strategies based on predicted dynamic changes in the VUT's driving behavior, allowing for continuous adversarial interaction according to the testing difficulty requirements. In the risk scenario exposure rate method, real-world interaction data is used to generate test scenarios with varying risk levels. By applying the importance sampling theorem, the probability of low probability, high-risk scenarios is increased, enabling the rapid generation of high-risk adversarial test scenarios for accelerated closed-course testing.

The interaction process is dynamic and continuous. To realistically showcase the testing process, we have also made some real-world test video clips available online. Refer to the supplementary videos:

1) Video1\_ScenarioA.mp4;

2) Video2\_ScenarioB.mp4.
\section{Conclusion and future research}\label{sec:Conclusion and future research}
This study presents Real-world Troublemaker, a novel track testing framework for automated driving systems in safety-critical interaction scenarios. It addresses the limitations of traditional track testing by enabling: (i) fully unmanned, automated testing with a cloud-controlled object target framework, (ii) precise reproduction of interaction scenarios, (iii) enhanced scenario diversity, and (iv) increased maneuvering challenges, and risk exposure frequency. The effectiveness of the Troublemaker system was validated through real-world experiments, demonstrating a 65.2\% improvement in scene reproduction accuracy, a 9.2 times increase in interaction strategy diversity, and a 3.5 times enhancement in high-risk scenario exposure.

Future work will refine the proposal function in importance sampling to dynamically adjust test difficulty, enabling iterative tests tailored to ADS performance. We also plan to extend the framework to handle more complex scenarios, including mixed human-vehicle traffic, to further validate its robustness and scalability in real-world applications.

\bibliographystyle{ieeetr}
\bibliography{ref}

\end{document}